\documentclass{sig-alternate-05-2015}

\DeclareMathOperator*{\argmin}{arg\,min}
\usepackage{multirow}
\usepackage[ruled,lined]{algorithm2e}
\usepackage{url}
\usepackage{hyperref}
\begin{document}






%

\title{Exploiting Multi-modal Curriculum in Noisy Web Data for Large-scale Concept Learning}
%
%
%
%
%
\pdfinfo{
/Title (Exploiting Multi-modal Curriculum in Noisy Web Data for Large-scale Concept Learning)
/Author (Junwei Liang, Lu Jiang, Deyu Meng, Alexander Hauptmann) }

\numberofauthors{4} 
%
\author{
%
%
Junwei Liang$^{1}$, Lu Jiang$^{1}$, Deyu Meng$^{2}$, Alexander Hauptmann$^{1}$ \\
$^1$School of Computer Science, Carnegie Mellon University, PA, USA\\
$^2$School of Mathematics and Statistics, Xi'an Jiaotong University, P. R. China.\\
\{junweil, lujiang, alex\}@cs.cmu.edu, dymeng@mail.xjtu.edu.cn.
}

\maketitle
\begin{abstract}
Learning video concept detectors automatically from the big but noisy web data with no additional manual annotations is a novel but challenging area in the multimedia and the machine learning community. A considerable amount of videos on the web are associated with rich but noisy contextual information, such as the title, which provides weak annotations or labels about the video content. To leverage the big noisy web labels, this paper proposes a novel method called WEbly-Labeled Learning (WELL), which is established on the state-of-the-art machine learning algorithm inspired by the learning process of human. WELL introduces a number of novel multi-modal approaches to incorporate meaningful prior knowledge called curriculum from the noisy web videos. To investigate this problem, we empirically study the curriculum constructed from the multi-modal features of the videos collected from YouTube and Flickr. The efficacy and the scalability of WELL have been extensively demonstrated on two public benchmarks, including the largest multimedia dataset and the largest manually-labeled video set. The comprehensive experimental results demonstrate that WELL outperforms state-of-the-art studies by a statically significant margin on learning concepts from noisy web video data. In addition, the results also verify that WELL is robust to the level of noisiness in the video data. Notably, WELL trained on sufficient noisy web labels is able to achieve a comparable accuracy to supervised learning methods trained on the clean manually-labeled data.
\end{abstract}

%
%



%
%

%
%

%
%



%
%

%
%


\keywords{Video Understanding, Prior Knowledge, Web Label, Big Data, Webly-supervised Learning}

\begin{figure}[!t]
	\centering
		\includegraphics[width=0.47\textwidth]{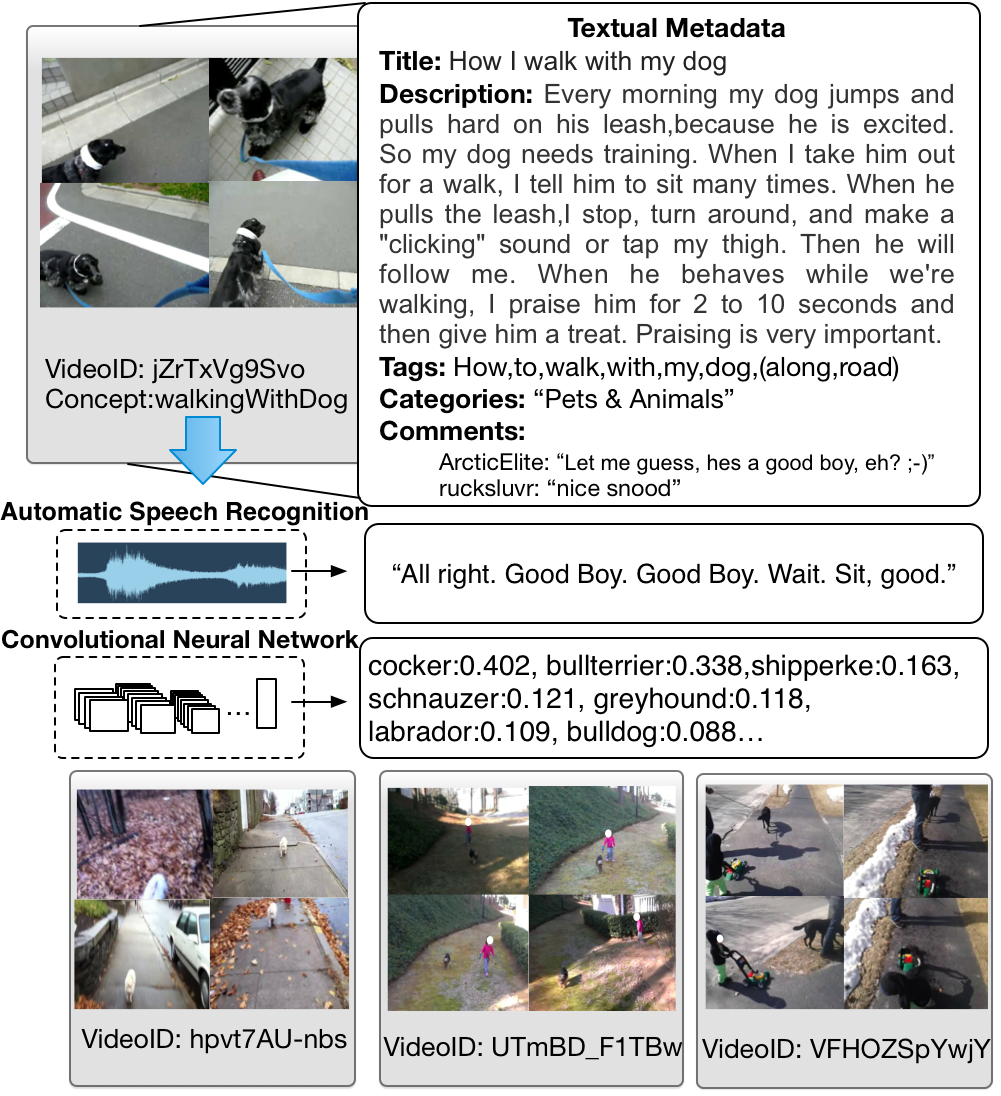}
	\caption{Multi-modal prior knowledge from web video. We investigate exploiting multi-modal information from the web videos without any manual effort to build concept detectors.}
	\label{cur-example}
	\vspace{-5mm}
\end{figure}

\section{Introduction}

Nowadays, millions of videos are being uploaded to the Internet every day. These videos capture all aspects of multimedia content about their uploader's daily life. These explosively growing user generated content videos online are becoming an crucial source of video data. Automatically categorizing videos into concepts, such as people actions, objects, etc., has become an important research topic. Recently many work have been proposed to tackle with building concept detectors both in image domain and video domain \cite{deng2009imagenet,liang2015towards,tang2012shifting,karpathy2014large,jiang2015exploiting}. However, the need for manual labels by human annotators has become one of the major important limitations for large-scale concept learning. It is even more so in video domain, since training concept detectors on videos is more challenging than on still images.
Many image datasets such as ImageNet~\cite{deng2009imagenet}, CIFAR~\cite{krizhevsky2009learning}, PASCAL VOC\cite{everingham2010pascal}, MS COCO~\cite{lin2014microsoft} and Caltech~\cite{fei2006one} have been collected and manually labeled. In video domain, some largest datasets such as UCF-101\cite{soomro2012ucf101}, MCG-WEBV~\cite{cao2009mcg}, TRECVID MED~\cite{over2014trecvid} and FCVID~\cite{jiang2015exploiting} are popular benchmark datasets for video classification. 
Collecting such datasets requires a large amount of human effort that can take thousands of man hours. In addition, manually labeling video requires playing back the video, which is more time consuming and expensive than labeling still images. As a result, the largest labeled video collection, FCVID~\cite{jiang2015exploiting}, only contains about 0.09 million labels with 239 concept classes, much less than the 14 million labels with over 20,000 classes in the image collection ImageNet~\cite{deng2009imagenet}.

Many state-of-the-art models in visual classification are based on the neural networks~\cite{jiang2015exploiting,karpathy2014large,varadarajan2015efficient}. As the architecture gets deeper, the neural network would need more data to train in order to get better performance. However, more data needs more human supervision which are more expensive to acquire in the video domain. 


Videos are available on the web and contain rich contextual information with a weak annotation about their content, such as their titles, descriptions and surrounding text. These webly-labeled data are orders of magnitude larger than that of any manually-labeled collections. Moreover, automatically extracted features from multiple modalities such as existing still image classification models, automatic speech recognition and optical character recognition tools can be useful additional information for the content of the video. 
Figure \ref{cur-example} shows an example of webly-labeled video for walking with a dog. As we see, the textual metadata we get from the web videos contain useful but very noisy information. The multi-modal prior information we get is correlated across modalities, as the image classification results and speech transcript show high probability of dog appearance, while the textual metadata indicates the same content. Some of the videos (about 20\% in the FCVID dataset) have very little textual metadata and we can only obtain web labels via other modalities.
To address the problem of learning detectors from the big web data, in this paper, we utilize multi-modal information to harness prior knowledge from the web video without any manual efforts.
 
Existing methods on learning from noisy webly-labeled data has mainly focused on the image domain \cite{fergus2005learning,li2010optimol,bergamo2010exploiting,chen2015webly}. Existing studies demonstrated promising results in this direction. However, these methods are primarily based on some heuristic methods. It is not clear what objective is being optimized and where or even whether the learning process will converge. Moreover, these methods only utilize a single text modality in the image domain. It is unclear how to exploit the multi-modal prior knowledge for concept learning from the rich context of Internet video.

To utilize the large amount of webly-labeled video data for concept learning, we propose a learning framework called \textbf{WEbly-Labeled Learning (WELL)}. It is established on the theories called \textit{curriculum learning}~\cite{bengio2009curriculum} and \textit{self-paced learning}~\cite{kumar2010self}. The learning framework is motivated by human learning, where people generally start learning easier aspects of a concept, and then gradually take more complex examples into the learning process\cite{bengio2009curriculum,kumar2010self,jiang2015self}. 
Following this idea, WELL learns a concept detector iteratively from first using a few samples with more confident labels (more related to the concept), then gradually incorporate more video samples with noisier labels. The algorithm combines the prior knowledge, called learning curriculum, extracted from the webly-labeled data with the dynamic information learned from the statistical model (self-paced) to determine which video samples to learn in the next iteration. This idea of easy-to-hard learning paradigm has been adopted for learning in noisy web data~\cite{jiang2014easy,chen2015webly,kumar2010self} and has been proved to be efficient to deal with noise and outliers. 
Our proposed method generalizes such learning paradigm using a clear objective function. It is proved to be convex and is a also general framework that can incorporate state-of-the-art deep learning methods to learn robust detectors from noisy data. Our framework fundamentally changes self-paced learning and allows learning for video concept detectors at unlimited scale.

\begin{figure*}
	\centering
		\includegraphics[width=0.9\textwidth,height=1.5in]{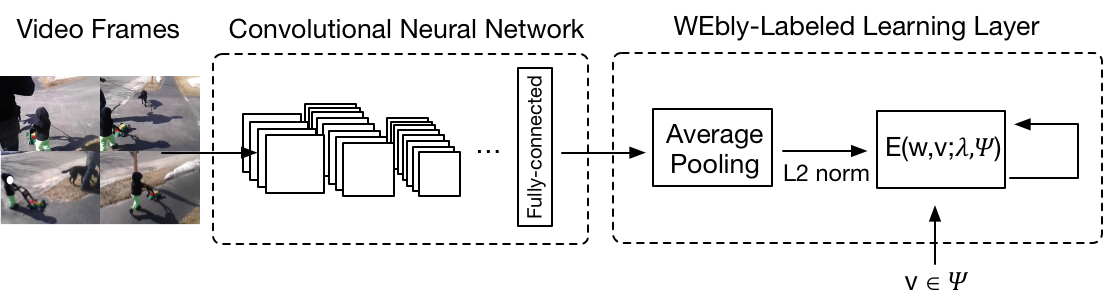}
	\vspace{-6mm}
	\caption{\textbf{Learning architecture of WEbly-Labeled Learning (WELL)}}
	\label{well-cnn}
\end{figure*}

Figure~\ref{well-cnn} shows the architecture of the proposed method. We extract keyframe-level convolutional neural network features and feed them into WELL layer with average pooling and iterative learning process. We have also tried using other video features such as motion features and audio MFCC features.

Our contributions are threefold. First, we address the problem of learning robust video concept detectors from noisy web data through a general framework with solid theoretical justifications. We show that WELL not only outperforms state-of-the-art learning methods on noisy labels, but also, notably, achieves comparable results with state-of-the-art models trained using manual annotation on one of the largest video dataset. Second, we provide detailed comparison of different approaches to exploit multi-modal curriculum from noisy labels and verify that our method is robust against certain level of noisiness in the video data. Finally, the efficacy and the scalability have been empirically demonstrated on two public benchmarks, including by far the largest manually-labeled video set called FCVID~\cite{jiang2015exploiting} and the largest multimedia dataset called YFCC100M~\cite{thomee2015yfcc100m}. The promising results suggest that detectors trained on sufficient webly-labeled videos may outperform detectors trained on any existing manually-labeled datasets. 

\section{Related Work}

\textbf{Curriculum and Self-paced Learning}: Recently a learning paradigm called~\textit{curriculum learning} (CL) was proposed by Bengio et al., in which a model is learned by gradually incorporating from easy to complex samples in training so as to increase the entropy of training samples~\cite{bengio2009curriculum}. A curriculum determines a sequence of training samples and is often derived by predetermined heuristics in particular problems. 
For example, Chen et al. designed a curriculum where images with clean backgrounds are learned before the images with noisy backgrounds~\cite{chen2015webly} , i.e. their method first builds a feature representation by a Convolutional Neural Network (CNN) on images with clean background and then they fine tune the models on images with noisy background. 
In~\cite{spitkovsky2009baby}, the authors approached grammar induction, where the curriculum is derived in terms of the length of a sentence. Because the number of possible solutions grows exponentially with the length of the sentence, and short sentences are easier and thus should be learn earlier.

The heuristic knowledge in a problem often proves to be useful. However, the curriculum design may lead to inconsistency between the fixed curriculum and the dynamically learned models. That is, the curriculum is predetermined a prior and cannot be adjusted accordingly, taking into account the feedback about the learner. To alleviate the issue of CL, Kumar et al. designed a learning paradigm, called \emph{self-paced learning} (SPL)~\cite{kumar2010self}. SPL embeds curriculum design as a regularizer into the learning objective. Compared with CL, SPL exhibits two advantages: first, it jointly optimizes the learning objective with the curriculum, and thus the curriculum and the learned model are consistent under the same optimization problem; second, the learning is controlled by a regularizer which is independent of the loss function in specific problems. This theory has been successfully applied to various applications, such as matrix factorization~\cite{zhao2015self}, action/event detection~\cite{jiang2014self}, domain adaption~\cite{tang2012shifting}, tracking~\cite{supancic2013self} and segmentation~\cite{kumar2011learning}, reranking~\cite{jiang2014easy}, etc.

\textbf{Learning Detectors in Web Data}: Many recent studies have been proposed to utilize the large amount of noisy data from the Internet. For example, \cite{mitchell2015never} proposed a Never-Ending Language Learning (NELL) paradigm and built adaptive learners that makes use of the web data by learning different types of knowledge and beliefs continuously. Such learning process is mostly self-supervised, and previously learned knowledge enables learning further types of knowledge.

In the image domain, existing methods try to tackle the problem of constructing qualified training sets based on the search results of text or image search engines~\cite{fergus2005learning,li2007optimol,chen2013neil,li2014exploiting,divvala2014learning,liang2015towards}. 
For example, \cite{fergus2005learning} extended the probabilistic Latent Semantic Analysis in visual domain and learned object categories using results from image search engines. \cite{li2007optimol} proposed an incremental learning paradigm that initialized from a few seed images and repeatedly trained models to refine the collected image dataset from the Internet.
NEIL~\cite{chen2013neil} followed the idea of NELL and learned from web images to form a large collection of concept detectors iteratively via a semi-supervised fashion. By combining the classifiers and the inter-concept relationships it learned, NEIL can be used for scene classification and object detection task.
\cite{li2014exploiting} tried to learn robust classifiers by considering the noisy textual information accompanied with web images. However, the down side is that the portion of the true positive samples has to be determined via prior knowledge, where in fact it is not accurate to assume the same number of true positive samples for any targeted concept.
\cite{divvala2014learning} introduced a webly-supervised visual concept learning method that automatically learns large amount of models for a wide range of variations within visual concepts. They discovered concept variances through vocabulary of online books, and then downloaded images based on text-search from the web to train object detection and localization models.
\cite{liang2015towards} presented a weakly-supervised method called Baby Learning for object detection from a few training images and videos. They first embed the prior knowledge into a pre-trained CNN. When given very few samples for a new concept, a simple detector is constructed to discover much more training instances from the online weakly labeled videos. As more training samples are selected, the concept detector keeps refining until a mature detector is formed. 
Another recent work in image domain \cite{chen2015webly} proposed a webly supervised learning of Convolutional Neural Network. They utilized easy images from search engine like Google to bootstrap a first-stage network and then used noisier images from photo-sharing websites like Flickr to train an enhanced model.  

In video domain, only few studies \cite{duan2012visual,han2015fast,varadarajan2015efficient} have been proposed for noisy data learning since training robust video concept detectors is more challenging than the problem in the image domain.
\cite{duan2012visual} tackled visual event detection problem by using SVM based domain adaptation method in web video data.
\cite{han2015fast} described a fast automatic video retrieval method using web images. Given a targeted concept, compact representations of web images obtained from search engines like Google, Flickr are calculated and matched to compact features of videos. Such method can be utilized without any pre-defined concepts.
\cite{varadarajan2015efficient} discussed a method that exploits the YouTube API to train large scale video concept detectors on YouTube. The method utilized a calibration process and hard negative mining to train a second order mixture of experts model in order to discover correlations within the labels. 

Most of the existing methods are heuristic approaches as it is unclear what objective is being optimizing on the noisy data. Moreover, results obtained from the web search results is just one approach to acquire prior knowledge or curriculum. To the best of our knowledge, there have been no systematical studies on exploiting the multi-modal prior knowledge in video concept learning on noisy data. Since search engine algorithm is changing rapidly, it is unclear that how noisy the web labels are and how the level of noisiness in the data will affect performance. In this paper, we proposed a theoretically justified method with clear framework for curriculum constructing and model learning. We also empirically demonstrate its superior performance over representative existing methods and systemically verify that WELL is robust against the level of noisiness of the video data.

\vspace{-1mm}
\section{WEbly-Labeled Learning (WELL)}
\subsection{Problem Description}
In this paper, following~\cite{varadarajan2015efficient}, we consider a concept detector as a classifier and our goal is to train concept detectors from webly-labeled video data without any manual labeling effort. Given a noisy web video training set and a target concept set, we do not assume any distribution of the noise. Formally, we represent the training set as $\mathcal{D} = \{(\mathbf{x}_i,\mathbf{z}_i,\mathbf{\tilde{y}}_i)\}_{i=1}^n$
where $\mathbf{x}_i \in \mathbb{R}^m$ denotes the feature for the $i^{th}$ observed sample, and $\mathbf{z}_i$ represents its noisy web label, which generally means the prior knowledge we can get from the web without additional human effort that not only includes textual information provided by the uploaders in the video metadata but also includes prior knowledge from other modalities using existing tools like pre-trained Convolutional Neural Network image detector \cite{chatfield2014return}, Automatic Speech Recognition \cite{povey2011kaldi} and Optical character recognition \cite{smith2007overview}. The $\mathbf{\tilde{y}}_i \subset \mathcal{Y}$ is the inferred concept label set for the $i^{th}$ observed sample based on its noisy web label, and $\mathcal{Y}$ denotes the full set of target concepts. In our experiment, to simplify the problem, we apply our method on binary classification and infer binary labels $\tilde{y}_i$ from the noisy web labels.
The noisy web labels can be used to automatically infer concept labels by matching the concept name to the video textual metadata. For example, a video may be inferred to the concept label ``cat'' as its textual title contains cat. \cite{varadarajan2015efficient} utilizes the YouTube topic API, which is derived from the textual metadata, to automatically get concept labels for videos. The web labels are quite noisy as the webly-labeled concepts may not present in the video content whereas the concepts not in the web label may well appear.

\subsection{Model and Algorithm}
\subsubsection{Objective Function}
To leverage the noisy web labels in a principled way, we propose WEbly-Labeled Learning (WELL). Formally, given a training set $\mathcal{D}$ as described before,  Let $L(\tilde{y}_i,g(\mathbf{x}_i,\mathbf{w}))$, or $\ell_i$ for short, denote the loss function which calculates the cost between the inferred label $\tilde{y}_i$ and the estimated label $g(\mathbf{x}_i,\mathbf{w})$. Here $\mathbf{w}$ represents the model parameter inside the decision function $g$. 
For example, in our paper, $\mathbf{w}$ represents the weight parameters in the Convolutional Neural Network (CNN) and the Support Vector Machine (SVM).
Our objective function is to jointly learn the model parameter $\mathbf{w}$ and the latent weight variable $\mathbf{v}= [v_1,\cdots,v_n]^T$ by:
\vspace{-2mm}
\begin{equation}
\label{eq:spcl_obj}
\begin{split}
\!\min_{\mathbf{w},\mathbf{v}\in \lbrack 0,\!1]^{n}}\!\!\mathbb{E}(\mathbf{w},\!\mathbf{v}\!;\lambda,\!\Psi\!)
\!=\! \sum_{i=1}^n v_i L(\tilde{y}_i,\!g(\mathbf{x}_i,\!\mathbf{w})) \!+\! f(\mathbf{v};\! \lambda), \\ \text{  subject to } \mathbf{v} \in \Psi
\end{split}
\end{equation}
where $\mathbf{v=[}v_{1},v_{2},\cdots ,v_{n}\mathbf{]}^{T}$ denote the
latent weight variables reflecting the inferred labels' confidence. The weights determine a learning sequence of samples, where samples with greater weights tend to be learned earlier. 
Our goal is to assign greater weights to the samples with more confident labels whereas smaller or zero weights to the samples with noisy labels. To this end, we employ the self-paced regularizer $f$, which controls the learning process of the model. 
We consider the linear regularizer Eq.~\eqref{eq:linear_scheme} proposed in~\cite{jiang2015self}:
\begin{equation}
\label{eq:linear_scheme}
f(\mathbf{v};\lambda) =  \frac{1}{2} \lambda \sum_{i=1}^n  ( v_i^2 - 2v_i ).
\vspace{-1mm}
\end{equation}

Generally, a self-paced regularizer determines the scheme for penalizing the latent weight variables. Physically it resembles the learning schemes human used in understanding new concepts. The linear scheme corresponds to a prudent strategy, which linearly penalizes the samples that are different to what the model has already learned (see Eq.~\eqref{eq:linear_closedform}). 
The hyper-parameter $\lambda$ $(\lambda > 0)$ is called ``model age'', which controls the pace at which the model learns new samples. When $\lambda$ is small only samples of with small loss will be considered. As $\lambda$ grows, more samples with larger loss will be gradually appended to train a ``mature'' mode.

$\Psi$ in Eq.~\eqref{eq:spcl_obj} is a curriculum region derived from noisy web labels $\mathbf{z}$ that incorporates the prior knowledge extracted from the webly-labeled data as a convex feasible region for the weight variables. The shape of the region weakly implies a prior learning sequence of samples, where the expected values for favored samples are larger. The curriculum region can be derived in a variety of ways that make use of different modalities. We will discuss this topic in details in following section. A straightforward approach is by counting the term frequency in the video's textual metadata. That is, for example, the chance of a video containing the concept ``cat'' become higher when it has more word ``cat'' in its title, description or tags. 

Eq.~\eqref{eq:spcl_obj} represents a concise and general optimization model ~\cite{jiang2015self}. It combines the prior knowledge extracted from the noisy webly-labeled data (as the curriculum region) and the information dynamically learned during the training (via the self-paced regularizer). Intuitively, the prior knowledge serves as an instructor providing a guidance on learning the latent weights, but it leaves certain freedom for the model (the student) to adjust the actual weights according to its learning pace. Experimental results in Section~\ref{sec:experiments} demonstrate the learning paradigm can better overcome the noisy labels than heuristic approaches. Figure \ref{well-cnn} shows the learning process of our method.

Following~\cite{kumar2010self,jiang2015self}, we employ the alternative convex search algorithm to solve Eq.~\eqref{eq:spcl_obj}. Algorithm~\ref{alg:overall} takes the input of a curriculum region, an instantiated self-paced regularizer and a step size parameter; it outputs an optimal model parameter $\mathbf{w}$. First of all, it initializes the latent weight variables in the feasible region. Then it alternates between two steps until it finally converges: Step 3 learns the optimal model parameter with the fixed and most recent $\mathbf{v}^*$; Step 5 learns the optimal weight variables with the fixed $\mathbf{w}^*$. In the beginning, the model ``age'' is gradually increased so that more noisy samples will be gradually incorporated in the training. Step 3 can be conveniently implemented by existing off-the-shelf supervised learning methods such as the back propagation. Gradient-based methods can be used to solve the convex optimization problem in Step 4. According to~\cite{gorski2007biconvex}, the alternative search in Algorithm~\ref{alg:overall} converges as the objective function is monotonically decreasing and is bounded from below.

\setlength{\textfloatsep}{1pt}
\vspace{-3mm}
\IncMargin{1em}
\begin{algorithm}
\SetKwData{Left}{left}\SetKwData{This}{this}\SetKwData{Up}{up}
\SetKwInOut{Input}{input}\SetKwInOut{Output}{output}
\LinesNumbered
\Input{Input dataset $\mathcal{D}$, curriculum region $\Psi$, self-paced function $f$ and a step size $\mu$}
\Output{Model parameter $\mathbf{w}$}
\BlankLine
Initialize $\mathbf{v}^*$, $\lambda$ in the curriculum region\;
\While{not converged} {
Update $\mathbf{w}^* = \arg\min_{\mathbf{w}} \mathbb{E}(\mathbf{w},\mathbf{v}^*;\lambda, \Psi)$\;
Update $\mathbf{v}^* = \arg\min_{\mathbf{v}} \mathbb{E}(\mathbf{w}^*,\mathbf{v}; \lambda, \Psi)$\;
\lIf{$\lambda$ is small}{increase $\lambda$ by the step size $\mu$}
}
\Return $\mathbf{w}^*$
\caption{\label{alg:overall} WEbly-Labeled Learning (WELL).}
\end{algorithm}
\DecMargin{1em}
\vspace{-3mm}

At an early age when $\lambda$ is small, Step 4 in Algorithm~\ref{alg:overall} has an evident suppressing effect over noisy samples that have greater loss to the already learned model. For example, with a fixed $\mathbf{w}$, the unconstrained close-formed solution for the regularizer in Eq.~\eqref{eq:linear_scheme} equals
\begin{equation}
\label{eq:linear_closedform}
v_i^* =\begin{cases}
-\frac{1}{\lambda} \ell_i +1 &  \ell_{i} < \lambda\\
0 & \ell_{i} \ge \lambda
\end{cases},
\end{equation}
where $v_i$ represents the $i$th element in the optimal solution $\mathbf{v}^* = [v_1^*, \cdots, v_n^*]^T$. Eq.~\eqref{eq:linear_closedform} called linear regularizer indicates the latent weight is proportional to the negative sample loss, and the sample whose loss is greater or equals to $\lambda$ will have zero weights and thus will not affect the training of the next model. As the model age grows, the hyper-parameter $\lambda$ increases, and more noisy samples will be used into training. The prior knowledge embedded in the curriculum region $\Psi$ is useful as it suggests a learning sequence of samples for the ``immature'' model. \cite{meng2015objective} theoretically proves that the iterative learning process is identical to optimizing a robust loss function on the noisy data.

If we keep increasing $\lambda$, the model will ultimately use every sample in the noisy data, which is undesirable as the labels of some noisy samples are bound to be incorrect. To this end, we stop increasing the age $\lambda$ after about a certain number of iterations (early stopping). The exact stopping iteration for each detector is automatically tuned in terms of its performance on a small validation set.

\begin{figure*}
	\centering
		\includegraphics[width=0.8\textwidth,height=0.3\textheight]{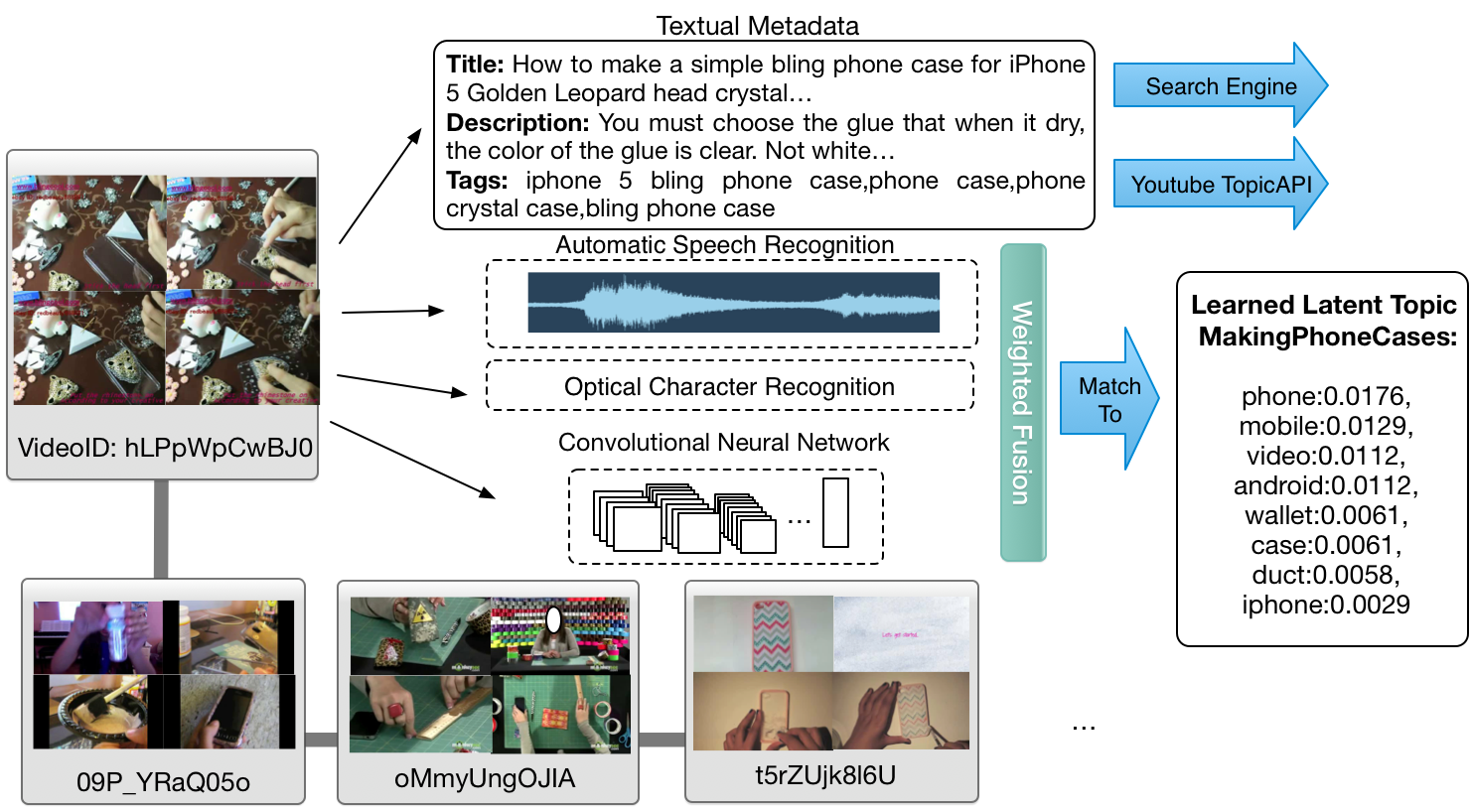}
	\caption{Curriculum Extraction Example. We automatically extract information using meaningful prior knowledge from several modalities and fuse them to get curriculum for WELL. Our method makes use of text, speech, visual cues while common methods like search engine only extract from textual information.}
	\label{curriculum_example}
\end{figure*}

\subsubsection{Model Details}\label{sec:partial_curriculum}
In this section we discuss further details of the curriculum region $\Psi$ and the self-paced regularizer $f(\mathbf{v};\lambda)$.

$\Psi$ is a feasible region that embeds the prior knowledge extracted from the webly-labeled data. It physically corresponds to a convex search region for the latent weight variable. Given a set of training samples $\mathbf{X}=\{\mathbf{x}_{i}\}_{i=1}^{n}$, 
we utilize the partial-order curriculum which generalizes the total-order curriculum by incorporating the incomplete prior over groups of samples. Samples in the confident groups should be learned earlier than samples in the less confident groups. It imposes no prior over the samples within the same group nor the samples not in any group. Formally, we define a partial order relation $\preceq$ such that $x_i \preceq x_j$ indicates that the sample $x_i$ should be learned no later than $x_j$ ($i,j \in [1,n]$). Similarly given two sample subsets $\mathbf{X}_{a} \preceq \mathbf{X}_{b}$ denotes the samples in $\mathbf{X}_{a}$ should be learned no later than the samples in $\mathbf{X}_{b}$.

In our problem, we extract the partial-order curriculum in the webly-labeled data in the following ways: we only distinguish the training order for groups of samples. 
Information from different modalities of the web labels can be used for curriculum design. A straightforward way is to directly utilize the textual descriptions of the videos generated by the uploaders. We compare common ways to extract curriculum from web data for concept learning to the proposed novel method that utilize state-of-the-art topic modeling techniques in natural language processing. 
In the following methods (Exact \& Stem Matching, Word Embedding and Latent Topic with Word Embedding), we first extract bag-of-words features from different modalities and then match them using specific matching methods to the concept words. Each video will then come with a matching score to each concept. In our experiment, we divide the data into two partial-order curriculum groups, where the videos with matching scores larger than zero will be in one group while others will be in the other group.

\textbf{Exact \& Stem Matching} We build curriculum directly using exact word matching or stemmed word matching between the textual metadata of the noisy videos to the targeted concept names. 

\textbf{YouTubeTopicAPI} We directly utilize the YouTube topic API to search for videos that are related to the concept words. The topic API utilizes textual information of the uploaded videos to obtain related topics of the videos from Freebase.

\textbf{SearchEngine} We build curriculum using the search result from a text-based search engine. It is similar to related web-search based methods.

\textbf{Word Embedding} We use word embedding \cite{mikolov2013distributed} to match words in metadata to targeted concept words in order to deal with synonyms and related concepts. The word embedding is trained using Google News data.

\textbf{Latent Topic} We build curriculum based on the latent topic we learned from the noisy label. We incorporate Latent Dirichlet Allocation (LDA)~\cite{blei2003latent} to determine how each noisy labeled video is related to each target concept. The basic idea is that each web video consists of mixtures of topics (concepts), and each topic is characterized by a distribution of words.
Formally, given all the noisy information extracted from a web video and collected them as a document $\mathbf{d_i}$, which combines into a corpus $\mathbf{d}$, we have a target set of $\mathbf{k}$  topics, then the key inferential problem that we are going to solve is that of computing the posterior distribution of the latent topics given a corpus (how likely the videos are related to each target concept given the noisy information):
\begin{equation}
\label{eq:lda}
p(\theta,\mathbf{t}|\mathbf{d},\alpha,\beta) = \frac{p(\theta,\mathbf{t},\mathbf{d}|\alpha,\beta)}{p(\mathbf{d}|\alpha,\beta)}
\end{equation}
where $\theta$ is the topic distribution variable for the corpus, $\theta$ $\sim$ Dir($\alpha$), in which Dir($\alpha$) represents a uniform Dirichlet distribution with scaling parameter $\alpha$. The $\mathbf{t}$ is the topic assignment variable that indicates which topic each word belong to in the document.
$\beta$ is the Dirichlet prior on the per-topic word distribution, in which we impose asymmetric priors over the word distribution so that each learned topic will be seeded with particular words in our target concept. For example, a topic will be seeded with words "walk, dog" for the target concept "WalkingWithDog".
The parameter estimation in Eq~\eqref{eq:lda} can be done via Bayes methods. 
However, Eq.~\eqref{eq:lda} is intractable to compute since $p(\mathbf{d}|\alpha,\beta)$ is intractable due to the coupling between $\theta$ and $\beta$ \cite{blei2003latent}. To solve this problem, approximate inference algorithms are introduced and we use the online variational inference algorithm from \cite{hoffman2010online}. The true posterior is approximated by a simpler distribution:
\begin{equation}
\label{eq:lda-vi}
q(\theta,\mathbf{t}|\gamma,\phi) = q(\theta|\gamma)\prod_{i=1}^{N}{q(t_i|\phi_i)}
\end{equation}
where the Dirichlet parameter $\gamma$ and the multinomial parameters ($\phi_1,...,\phi_N$) are the free variational parameters. Thus the maximization problem is equivalent to minimizing the Kullback-Leibler(KL) divergence between $q(\theta,\mathbf{t}|\gamma,\phi)$ and the posterior $p(\theta,\mathbf{t}|\mathbf{d},\alpha,\beta)$ \cite{blei2003latent}. The optimization problem can then be solve using Expectation-Maximization algorithm \cite{hoffman2010online}. The estimated parameters $\theta,t$ in Eq.~\eqref{eq:lda} is then used for constructing curriculum region in Eq.~\eqref{eq:spcl_obj}.

\textbf{Latent Topic with Word Embedding (LT+WE)} We first learn latent topics using LDA to replace the concept words with a topic word distribution and then match the latent topic words to the web label's bag-of-words features by using the word embeddings. We compare this method to the others using only textual information from the web videos. 
We also use this method to get curriculum from other modalities such as Automatic Speech Recognition (ASR) \cite{povey2011kaldi}, Optical Character Recognition (OCR) \cite{smith2007overview} and basic image detector pre-trained on still images~\cite{ILSVRC15} (in this paper we use VGG net~\cite{simonyan2014very}, extract keyframe-level image classification results and average them to get video-level results.). 
We extract bag-of-word features from them and combine them with linear weights. 
Detailed fusion experiments can be found in Section \ref{sec:experiments}. We empirically set OCR's weight to be small as the results are much noisier than other features.

Figure~\ref{curriculum_example} shows an example of the noisy web video data and how the curriculum is extracted with different methods. Our method can utilize information from different modalities while common methods like search engine only consider textual information.
We compare the performance of different ways of curriculum design by training detectors directly in Section 4.

The labels in webly-labeled data are much noisier than manually-labeled data, and as a result, we found that the learning is prone to overfitting the noisy labels. To address this issue, inspired by the dropout technique in deep learning~\cite{srivastava2014dropout}, we use a dropout strategy for webly-labeled learning~\cite{liang2016learning}. It is implemented in the self-paced regularizer discussed in Section 3. With the dropout, the regularizers become:
\begin{equation}
\label{eq:dropout}
\vspace{-2mm}
\begin{split}
r_i(p) \sim \text{Bernoulli}(p) + \epsilon, (0 < \epsilon \ll 1)\\
f(\mathbf{v};\lambda, p) =  \frac{1}{2} \lambda \sum_{i=1}^n  (\frac{1}{r_i} v_i^2 - 2v_i),
\end{split}
\vspace{-2mm}
\end{equation}
where $\mathbf{r}$ is a column vector of independent Bernoulli random variables with the probability $p$ of being 1. Each of the element equals the addition of $r_i$ and a small positive constant $\epsilon$. Denote $\mathbb{E}_{\mathbf{w}} = \sum_{i=1}^n v_i \ell_i + f(\mathbf{v};\lambda)$ as the objective with the fixed model parameters $\mathbf{w}$ without any constraint, and the optimal solution $\mathbf{v}^* = [v_1^*, \cdots, v_n^*]^T= \argmin_{\mathbf{v}\in[0,1]^n} \mathbb{E}_{\mathbf{w}}$. We have:
\begin{equation}
\vspace{-2mm}
\label{eq:linear_dropout}
\begin{split}
\mathbb{E}_{\mathbf{w}} = \sum_{i=1}^n \ell_i v_i + \lambda ( \frac{1}{2r_i} v_i^2 - v_i );\\
\frac{\partial \mathbb{E}_{\mathbf{w}}}{\partial v_i} =  \ell + \lambda v_i/r_i  -\lambda = 0;\\
\Rightarrow v_i^* =\begin{cases}
r_i(-\frac{1}{\lambda} \ell_i +1) &  \ell_{i} < \lambda\\
0 & \ell_{i} \ge \lambda
\end{cases}.
\end{split}
\vspace{-2mm}
\end{equation}
The dropout effect can be demonstrated in the closed-form solutions in Eq.~\eqref{eq:linear_dropout}: with the probability $1-p$, $v_i^*$ in both the equations approaches 0; with the probability $p$, $v_i^*$ approaches the solution of the plain regularizer discussed in Eq.~\eqref{eq:linear_scheme}. Recall the self-paced regularizer defines a scheme for learning samples. Eq.~\eqref{eq:linear_dropout} represent the new dropout learning scheme.

When the base learner is neural networks, the proposed dropout can be used combined with the classical dropout in~\cite{srivastava2014dropout}. The term dropout in this paper refers to dropping out samples in the iterative learning. By dropping out a sample, we drop out its update to the model parameter, which resembles the classical dropout used in neural networks. It operates on a more coarse-level which is useful for noisy data. When samples with incorrect noisy labels update a model, it will encourage the model to select more noisy labels. The dropout strategy prevents overfitting to noisy labels. It provides a way of combining many different sample subsets in different iterations in order to help avoid bad local minima. Experimental results substantiate this argument. In practice, we recommend setting two Bernoulli parameters for positive and negative samples on imbalanced data.

\section{Experiments}\label{sec:experiments}
In this section, we evaluate our method WELL for learning video detectors on noisy labeled data. The experiments are conducted on two major public benchmarks: FCVID and YFCC100M, where FCVID is by far one of the biggest manually annotated video dataset \cite{jiang2015exploiting}, and the YFCC100M dataset is the largest multimedia benchmark \cite{thomee2015yfcc100m}.

\subsection{Experimental Setup}
\textbf{Datasets, Features and Evaluation Metrics}
    Fudan-columbia Video Dataset (FCVID) contains 91,223 YouTube videos (4,232 hours) from 239 categories. It covers a wide range of concepts like activities, objects, scenes, sports, DIY, etc. Detailed descriptions of the benchmark can be found in~\cite{jiang2015exploiting}. Each video is manually labeled to one or more categories. In our experiments, we do not use the manual labels in training, but instead we automatically generate the web labels according to the concept name appearance in the video metadata. The manual labels are used only in testing to evaluate our and the baseline methods. Following~\cite{jiang2015exploiting}, the standard train/test split is used.
    The second set is YFCC100M \cite{thomee2015yfcc100m} which contains about 800,000 videos on Yahoo! Flickr with metadata such as the title, tags, the uploader, etc. There are no manual labels on this set and we automatically generate the curriculum from the metadata in a similar way. Since there are no annotations, we train the concept detectors on the most 101 frequent latent topics found in the metadata. 
    There are totally 47,397 webly labeled videos on the 101 concepts for training. 
    
    On FCVID, as the manual labels are available, the performance is evaluated in terms of the precision of the top 5 and 10 ranked videos (P@5 and P@10) and mean Average Precision (mAP) of 239 concepts. On YFCC100M, since there are no manual labels, for evaluation, we apply the detectors to a third public video collection called TRECVID MED which includes 32,000 Internet videos~\cite{over2014trecvid}. We apply the detectors trained on YFCC100M to the TRECVID videos and manually annotate the top 10 detected videos returned by each method for 101 concepts.

\textbf{Implementation Details}
    We build our method on top of a pre-trained convolutional neural network as the low-level features (VGG network \cite{simonyan2014very}). We extract the key-frame level features and create a video feature by the average pooling. The same features are used across different methods on each dataset. The concept detectors are trained based on a hinge loss cost function. Algorithm~\ref{alg:overall} is used to train the concept models iteratively and the $\lambda$ stops increasing after 100 iterations. We automatically generate curriculum labels based on the video metadata, ASR, OCR and VGG net 1,000 classification results using latent topic modeling with word embedding matching as shown in Section 3, and derive a partial-order curriculum using the method discussed in Section 3.2.2.

\textbf{Baselines}
    The proposed method is compared against the following five baseline methods which cover both the classical and the recent representative learning algorithms on webly-labeled data. 
    \textit{BatchTrain} trains a single SVM model using all samples in the multi-modal curriculum built as described in section 3.2.2 LT+WE.
    \textit{Self-Paced Learning (SPL)} is a classical method where the curriculum is generated by the learner itself~\cite{kumar2010self}. 
    \textit{BabyLearning} is a recent method that simulates baby learning by starting with few training samples and fine-tuning using more weakly labeled videos crawled from the search engine \cite{liang2015towards}. 
    \textit{GoogleHNM} is a hard negative mining method proposed by Google \cite{varadarajan2015efficient}. It utilizes hard negative mining to train a second order mixture of experts model according to the video's YouTube topics. \textit{FastImage} \cite{han2015fast} is a video retrieval method that utilizes web images from search engine to match to the video with re-ranking. \textit{WELL} is the proposed method. The hyper-parameters of all methods including the baseline methods are tuned on the same validation set. On FCVID, the set is a standard development set with manual labels randomly selected from 10\% of the training set  (No training was done using ground truth labels) whereas on YFCC100M it is also a 10\% proportion of noisy training set.
    
\subsection{Experiments on FCVID}
    \textbf{Curriculum Comparison} As disscussed in Section 3.2.2, we compare different ways to build curriculum for noisy label learning. 
    Here we also compare their effectiveness by training concept detectors directly using the curriculum labels. The batch train model is used for all generated cirriculumn labels. 
    In Table \ref{exp-cur} we show the batch trained models' precision at 5, 10 and mean average precision on the test set of FCVID. 
    For LT+WE (Multi-modal), we extract curriculum from different modalities as shown in Section 3.2.2, and combine them using linear weights. The weights are hyper-parameters that are tuned on the validation set, and the optimal weights for textual metadata, ASR, image classification and OCR results are 1.0, 0.5, 0.5 and 0.05, respectively.
    Results show that the curriculum generated by combining latent topic modeling and word embedding using multi-modal prior knowledge is the most accurate, which indicates our claim of exploiting multi-modal information is beneficial, and we use this method in WELL for the rest of the experiments.

\begin{table}[]
\centering
\caption{Curriculum BatchTrain Comparison}
\label{exp-cur}
\begin{tabular}{|l||c|c|c|}
\hline
Method            & P@5    & P@10   & mAP    \\ \hline \hline
ExactMatching             &   0.730     &   0.713     &      0.419  \\ 
StemMatching              & 0.782 & 0.763 & 0.469 \\ 
YouTubeTopicAPI &     0.587   &    0.563    &    0.315    \\ 
SearchEngine      &    0.723    &   0.713     & 0.413 \\ 
WordEmbedding          &     0.790   &     0.774   &     0.462   \\ 
LatentTopic          &   0.731    &     0.716   &    0.409   \\ 
LT+WE          &   0.804    &    0.795    &      0.473  \\ 
\textbf{LT+WE(Multi-modal)}               &   \textbf{0.838 }    &   \textbf{ 0.820 }   &     \textbf{ 0.486 } \\ \hline
\end{tabular}
\end{table}

    \textbf{Baseline Comparison} Table~\ref{exps-basline} compares the precision and mAP of different methods where the best results are highlighted. As we see, the proposed WELL significantly outperforms all baseline methods, with statistically significant difference at $p$-level of 0.05. Comparing WELL with SPL, the effect of curriculum learning and dropout makes a significant difference in terms of performance, which suggests the importance of prior knowledge and preventing over-fitting in webly learning. The promising experimental results substantiate the efficacy of the proposed method.
    
\begin{table}[ht]
\centering
\footnotesize

\caption{Baseline comparison on FCVID}

\label{exps-basline}

\begin{tabular}{|l||c|c|c|c|c|c|}
\hline
	   Method      & P@5 & P@10  & mAP \\ \hline \hline
	BatchTrain             & 0.838 & 0.820 & 0.486  \\
	FastImage~\cite{han2015fast} &- &- & 0.284\\
	SPL~\cite{kumar2011learning}           &  0.793 & 0.754  & 0.414    \\
    GoogleHNM~\cite{varadarajan2015efficient}      &  0.781 &  0.757 & 0.472  \\ 
	BabyLearning~\cite{liang2015towards}   & 0.834  & 0.817  & 0.496  \\ 
    \textbf{WELL}      &\textbf{0.918}&\textbf{0.906} & \textbf{0.615}  \\ \hline
	\end{tabular}
\end{table}

  \textbf{Robustness to Noise Comparison}
  In this comparison we manually control the noisiness of the curriculum in order to systematically verify how our methods would perform with respect to the noisiness within the web data. The experimental results indicate the robustness of our method towards noisy labels. To this end, we randomly select video samples with ground truth labels for each concept, so that the precision of the curriculum labels are set at 20\%, 40\%, 60\%, 80\%  and we fix the recall of all the labels. We then train WELL using such curriculum and test them on the FCVID testing set. We also compare WELL to three other methods with the same curriculum, among them \textit{GoogleHNM} is a recent method to train video concept detector with large-scale data. We exclude \textit{BabyLearning}, which relies on the returned results by the search engine, since in this experiment the curriculum is fixed . As shown in Table \ref{exp-noise}, as the noisiness of the curriculum grows (the precision drops), WELL maintains its performance while other methods drop significantly. Specifically, when the precision of the curriculum drops from 40\% to 20\%, other methods' mAP averagely drops 46.5\% while WELL's mAP only drops 19.1\% relatively. It shows that WELL is robust against different level of noise, which shows great potential in larger scale webly-labeled learning as the dataset gets bigger, the noisier it may become.
  
\begin{figure}[!ht]
	\centering
		\includegraphics[width=1.0\linewidth,height=55mm]{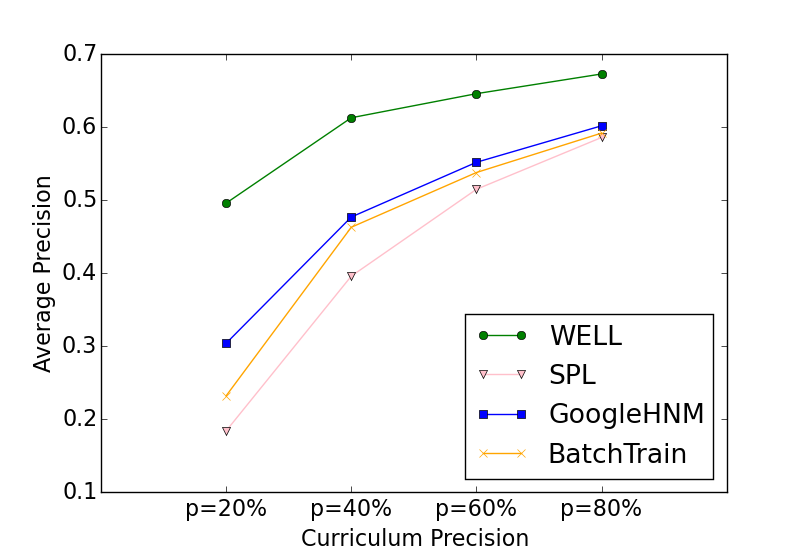}
	\caption{WELL performance with curriculum of different level of noisiness. p=k\% means the curriculum precision. The higher is k, the less noise is in the curriculum labels.}
	\label{well-noise}
\end{figure}

\begin{table}[ht]
\centering
\footnotesize
\caption{WELL performance with curriculum of different level of noisiness. $p$ represents the precision of the curriculum. }
\vspace{1mm}
\label{exp-noise}

\begin{tabular}{|l||c|c|c|c|}
\hline
     & p=20\% & p=40\% & p=60\% & p=80\% \\ \hline \hline
BatchTrain &     0.232   &   0.463     &    0.538    &   0.592    \\    
SPL &      0.184   &     0.396   &    0.515    &    0.586     \\ 
GoogleHNM & 0.304    &   0.477     &     0.552   &    0.602   \\ 
\textbf{WELL} &    \textbf{0.496}    &      \textbf{0.613}  &    \textbf{0.646}    &    \textbf{0.673}   \\ \hline
\end{tabular}
\end{table}
\begin{table}[ht]
\centering
\footnotesize
\caption{WELL performance with curriculum of different level of noisiness. $p$ represents the precision of the curriculum. }
\vspace{1mm}
\label{exp-noise}

\begin{tabular}{|l||c|c|c|c|}
\hline
     & p=80\% & p=60\% & p=40\% & p=20\% \\ \hline \hline
BatchTrain  &   0.592&    0.538&   0.463&     0.232               \\    
SPL           &    0.586&    0.515&     0.396&      0.184     \\ 
GoogleHNM             &    0.602&     0.552&   0.477& 0.304   \\ 
\textbf{WELL}           &    \textbf{0.673}&    \textbf{0.646}&      \textbf{0.613}&    \textbf{0.496}   \\ \hline
\end{tabular}
\end{table}
    \textbf{Ground-truth Training Comparison} 
    In this part, we also compare our method with the state-of-the-art method trained using ground truth labels on FCVID (rDNN)~\cite{jiang2015exploiting}. We compare WELL trained using the static CNN features, the standard features provided by the authors~\cite{jiang2015exploiting}, and we also compare WELL using the late (average) fusion with CNN, motion and audio MFCC features (WELL-MM) to the method that achieves the best result on FCVID trained using the same multi-modal features. WELL-MM uses CNN, Motion and MFCC features, which is the same set of features as rDNN-F~\cite{jiang2015exploiting}. Noted that the state-of-the-art method uses the ground truth labels to train models, which includes 42,223 videos with manual labels, while our proposed method uses none of the human annotation into training but still be able to outperform one of the state-of-the-art results.
    
 \begin{table}[ht]
\centering
\footnotesize

\caption{Ground-truth Training Comparison on FCVID. The methods with * are trained using human annotated labels. WELL-MM uses CNN, Motion and MFCC features, which is the same set of features as rDNN-F.}
\vspace{1mm}
\label{exps-sa}

\begin{tabular}{|l||c|c|c|c|c|c|}
\hline
	   Method      & P@5 & P@10  & mAP \\ \hline \hline
    WELL      &\textbf{0.918}&\textbf{0.906} & \textbf{0.615} \\ 
    Static CNN\cite{jiang2015exploiting}*      &-&- & 0.638  \\ 
    WELL-MM     &\textbf{0.930}&\textbf{0.918} & \textbf{0.697}  \\ 
   rDNN-F\cite{jiang2015exploiting}*   &-&- & 0.754  \\ \hline
	\end{tabular}
\end{table}

    \textbf{Noisy Dataset Size Comparison} To investigate the potential of concept learning on webly-labeled video data, we apply the methods on different sizes of subsets of the data. Specifically, we randomly split the FCVID training set into several subsets of 200, 500, 1,000, and 2,000 hours of videos, and train the models on each subset without using manual annotations. The models are then tested on the same test set. Table~\ref{exps-small} lists the average results of each type of subsets. As we see, the accuracy of WELL on webly-labeled data increases along with the growth of the size of noisy data while other webly learning methods' performance tend to be saturated.
    
Comparing to the methods trained using ground truth, In Table~\ref{exps-small}, WELL-MM trained using the whole dataset (2000h) outperforms rDNN-F(trained using manual labels) trained using around 1200h of data. And since the incremental performance increase of WELL-MM is close to linear, we conclude that with sufficient webly-labeled videos WELL-MM will be able to outperform the rDNN-F trained using 2000h of data, which is currently the largest manual labeled dataset.
    
    \vspace{-3mm}
\begin{figure}[!ht]
	\centering
		\includegraphics[width=1.05\linewidth,height=70mm]{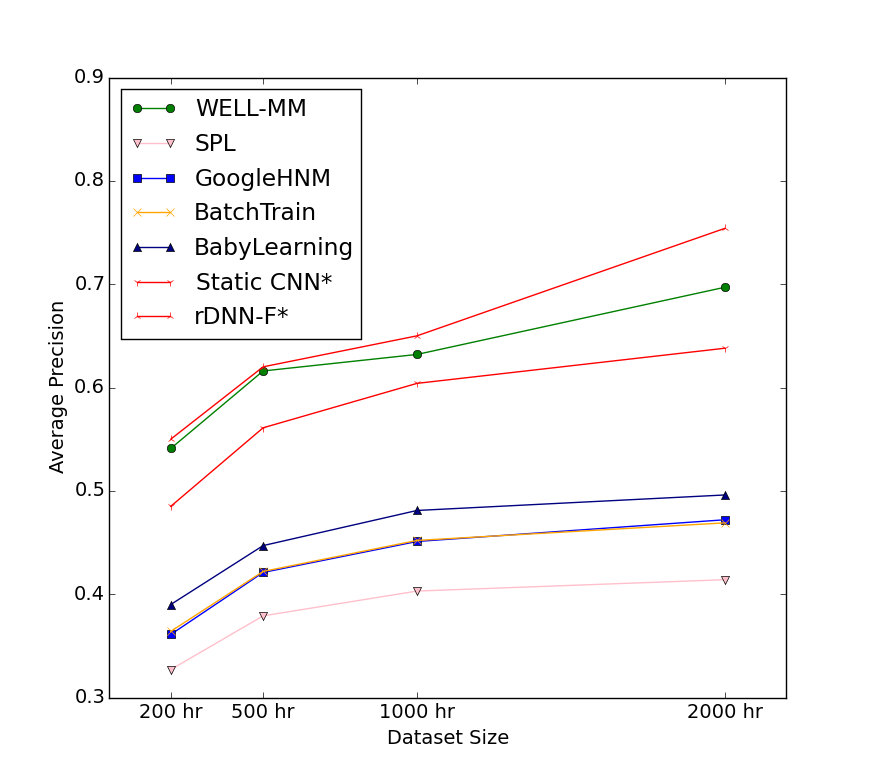}
	\vspace{-7mm}
	\caption{MAP comparison of models trained using web labels and ground-truth labels on different subsets of FCVID. The methods with * are trained using human annotated labels.}
	\label{well-dataset}
\end{figure}

    \begin{table}[ht]
\centering
\footnotesize
\caption{MAP comparison of models trained using web labels and ground-truth labels on different subsets of FCVID. The methods with * are trained using human annotated labels. Noted some of the numbers from \cite{jiang2015exploiting} are approximated from graphs.}
\vspace{1mm}
\label{exps-small}
\setlength{\tabcolsep}{4pt} 
\renewcommand{\arraystretch}{1.1} 
\begin{tabular}{|l||c|c|c|c|}
\hline
Dataset Size & 200h & 500h & 1000h & 2000h \\ \hline
BatchTrain     & 0.364    & 0.422     & 0.452    & 0.486    \\ 
	SPL~\cite{kumar2011learning}           & 0.327 &0.379 &0.403 &0.414 \\
    GoogleHNM~\cite{varadarajan2015efficient}     & 0.361   & 0.421  &0.451 &0.472 \\ 
	BabyLearning~\cite{liang2015towards}   &0.390   & 0.447    & 0.481    & 0.496 \\ 
\textbf{WELL-MM}     &  \textbf{0.541}  &  \textbf{0.616}    & \textbf{0.632}  & \textbf{0.697}   \\ \hline
Static CNN\cite{jiang2015exploiting}*        &     0.485     &   0.561        &    0.604      & 0.638 \\ 
rDNN-F\cite{jiang2015exploiting}* &   0.550    &   0.620       &      0.650    &  0.754\\ \hline 
\end{tabular}
\end{table}

\vspace{-3mm}

\subsection{Experiments on YFCC100M}
In the experiments on YFCC100M, we train 101 concept detectors on YFCC100M and test them on the TRECVID MED dataset which includes 32,000 Internet videos. Since there are no manual labels, to evaluate the performance, we manually annotate the top 10 videos in the test set and report their precisions in Table~\ref{exps-yfcc}. The MED evaluation is done by four annotators and the final results are averaged from all annotations. The Fleiss' Kappa value for these four annotators is 0.64.
A similar pattern can be observed where the comparisons substantiate the rationality of the proposed webly learning framework. Besides, the promising results on the largest multimedia set YFCC100M verify the scalability of the proposed method.

\vspace{-3mm}
\begin{table}[ht]
\centering
\footnotesize

\caption{Baseline comparison on YFCC100M}

\label{exps-yfcc}

\begin{tabular}{|l||c|c|c|c|c|c|}
\hline
Method      & P@3 & P@5   & P@10\\ \hline \hline
	BatchTrain              & 0.535  &  0.513 &  0.487 \\
	SPL~\cite{kumar2011learning}              &0.485  &   0.463 &  0.454\\
    GoogleHNM~\cite{varadarajan2015efficient}       &0.541  &  0.525 &  0.500 \\ 
	BabyLearning~\cite{liang2015towards}     &0.548  & 0.519 &  0.466 \\ 
	\textbf{WELL} &\textbf{0.667}    &   \textbf{0.663}   & \textbf{0.649}\\ \hline
	\end{tabular}
\end{table}
\begin{figure*}
	\centering
		\includegraphics[width=1.0\textwidth]{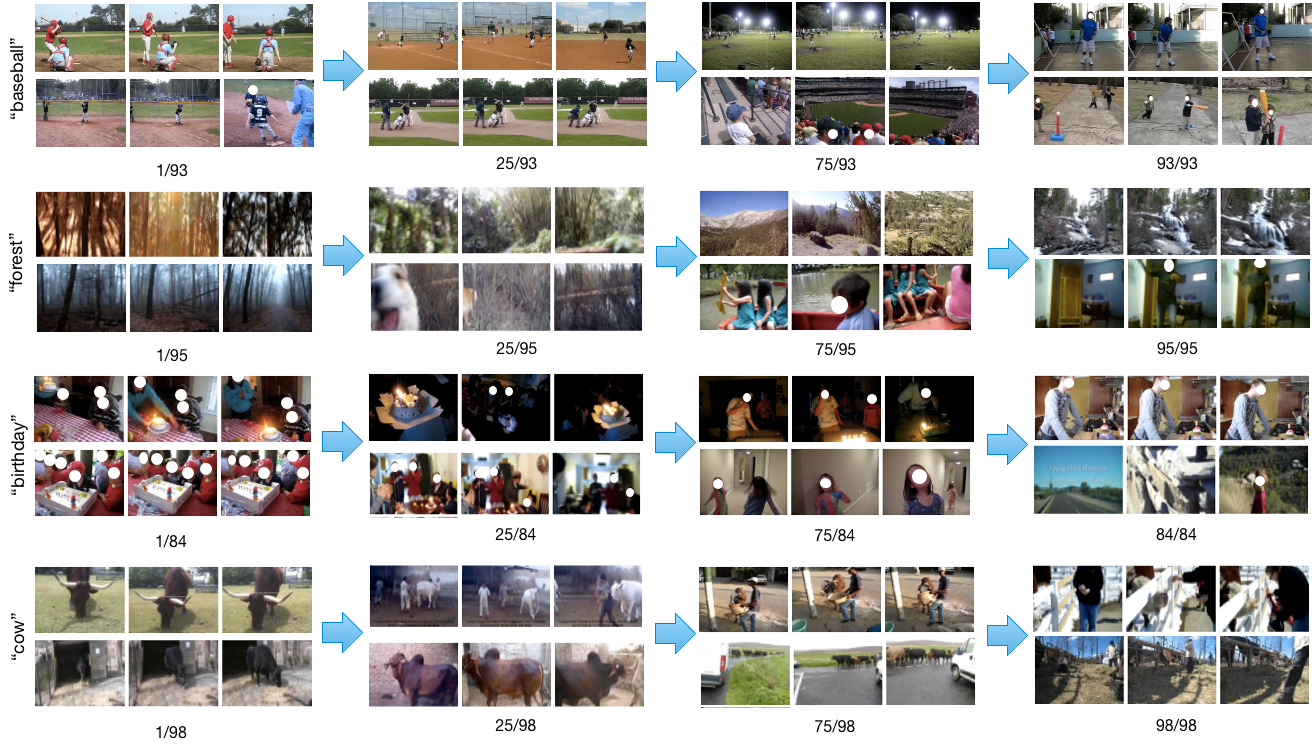}
		\vspace{-7mm}
	\caption{WELL's example picks for different iterations}
	\vspace{-5mm}
	\label{well-pick-all}
\end{figure*}
\vspace{-4mm}
\subsection{Time Complexity Comparison}
    The computation complexity of WELL is comparable to existing methods. For a single class, the complexity for our model and baseline models is $O(r \times n \times m)$, where $r$ is the number of iterations to converge, $n$ is the number of training samples and $m$ is the feature dimension. Theoretically, the complexity is comparable to the baseline models that have different $r$. In practice, on a 40 core-CPU machine, WELL and SPL takes 7 hours to converge (100 iterations) on FCVID with 239 concepts, whereas GoogleHNM and BabyLearning take around 5 hours. 
    In Table \ref{time}, we show the theoretical and actual run time for all methods.
    
\begin{table}[ht]
\centering
\caption{Runtime comparison across different methods. We report the time complexity on different method as well as their actual run time on FCVID (in hours).}
\label{time}
\begin{tabular}{|l||c|c|c|}
\hline
Method       & Complexity & FCVID(h) \\ \hline \hline
BatchTrain   &     $O(n \times m)$      &    2.0 \\
SPL          &      $O(r \times n \times m)$      &    7.0  \\
GoogleHNM    &  $O(r \times n \times m)$          &     5.0 \\
BabyLearning &  $O(r \times n \times m)$          &     5.0 \\
WELL         &      $O(r \times n \times m)$      &    7.0\\ \hline
\end{tabular}
\end{table}

\vspace{-3mm}
\subsection{Qualitative Analysis}
    In this section we show training examples of WELL. In Figure~\ref{well-pick-all}, we demonstrate the positive samples that WELL select at different stage of training the concept "baseball", "forest", "birthday" and "cow". 
    For the concept "baseball", at early stage (1/93, 25/93), WELL selects easier and clearer samples such as those camera directly pointing at the playground, while at later stage (75/93, 93/93) WELL starts to train with harder samples with different lighting conditions and untypical samples for the concept. 
    For the concept "birthday", as we see, at later stage of the training, complex samples for birthday event like a video with two girl singing birthday song (75/84) and a video of celebrating birthday during hiking (84/84) are included in the training.
    For the concept "forest", at the final iteration (95/95), a video of a man playing nunchaku is included, as the video title contains "Air Forester Nunchaku Freestyle" and it is included in the curriculum, which is reasonable as the curriculum is noisy and may contain false positive. Since WELL is able to leave outliers in later stage of the training, the affection of the false positives in the curriculum can be alleviated by early stopping. In our experiments, we stop increasing $\lambda$ after 100 iterations so in the final model, only a subset of samples are used.

\vspace{-2mm}
\section{Conclusions}

In this paper, we proposed a novel method called WELL for webly labeled video data learning. WELL extracts multi-modal informative knowledge from noisy weakly labeled video data from the web through a general framework with solid theoretical justifications. 
WELL achieves the best performance only using webly-labeled data on two major video datasets. The comprehensive experimental results demonstrate that WELL outperforms state-of-the-art studies by a statically significant margin on learning concepts from noisy web video data. In addition, the results also verify that WELL is robust to the level of noisiness in the video data. The result suggests that with more webly-labeled data, which is not hard to obtain, WELL can potentially outperform models trained on any existing manually-labeled data.


%
\vspace{-3mm}
\bibliographystyle{abbrv}
\linespread{0.9}    
\scriptsize
\bibliography{sigproc}  
%
%

\end{document}